\documentclass[numbers]{assets/matterlab}

\usepackage{microtype}
\usepackage{graphicx}
\usepackage{booktabs}
\usepackage{float}
\usepackage{xurl}
\usepackage{hyperref}
\usepackage{titlesec}
\usepackage{wrapfig}

\usepackage{amsmath}
\usepackage{amssymb}
\usepackage{mathtools}
\usepackage{amsthm}
\usepackage{bm}

\usepackage[noabbrev,nameinlink]{cleveref}

\usepackage{amsmath,amsfonts,bm}

\def\eqref#1{equation~\ref{#1}}

\def\1{\bm{1}}

\DeclareMathAlphabet{\mathsfit}{\encodingdefault}{\sfdefault}{m}{sl}
\SetMathAlphabet{\mathsfit}{bold}{\encodingdefault}{\sfdefault}{bx}{n}

\usepackage{orcidlink}
\usepackage{comment}
\usepackage[version=4]{mhchem}
\usepackage{longtable}
\usepackage{array}
\renewcommand{\arraystretch}{1.7}
\usepackage{multirow}           
\usepackage{amsfonts}           
\usepackage{nicefrac}
\usepackage{duckuments}         
\usepackage{thmtools,thm-restate}
\usepackage{enumitem}
\usepackage{xcolor,colortbl}
\usepackage{tikz}
\usepackage{tikz-cd}
\usepackage{caption}
\usepackage{subcaption}
\usepackage{listings}
\usepackage{gensymb}
\usepackage{textcomp} 
\usepackage[inkscapelatex=false]{svg}

\newtheorem{theorem}{Theorem}

\newtheorem{proposition}[theorem]{Proposition}

\newcommand{\addressCHEM}{Department of Chemistry, University of Toronto,  80 St. George St., Toronto, ON M5S 3H6, Canada}
\newcommand{\addressAC}{Acceleration Consortium, 700 University Ave., Toronto, ON M7A 2S4, Canada}
\newcommand{\addressCS}{Department of Computer Science, University of Toronto, 40 St George St., Toronto, ON M5S 2E4, Canada}
\newcommand{\addressVECTOR}{Vector Institute for Artificial Intelligence, W1140-108 College St., Schwartz Reisman Innovation Campus, Toronto, ON M5G 0C6, Canada}
\newcommand{\addressMSE}{Department of Materials Science \& Engineering, University of Toronto, 184 College St., Toronto, ON M5S 3E4, Canada}
\newcommand{\addressCHEMENG}{Department of Chemical Engineering \& Applied Chemistry, University of Toronto, 200 College St., Toronto, ON M5S 3E5, Canada}
\newcommand{\addressMEDICALSCI}{Institute of Medical Science, 1 King's College Circle, Medical Sciences Building, Room 2374, Toronto, ON M5S 1A8, Canada}
\newcommand{\addressCIFAR}{Canadian Institute for Advanced Research (CIFAR), 661 University Ave., Toronto,
ON M5G 1M1, Canada}
\newcommand{\addressNVIDIA}{NVIDIA, 431 King St W \#6th, Toronto, ON M5V 1K4, Canada}

\newcommand{\addressGT}{Georgia Institute of Technology, North Avenue, Atlanta, GA 30332, USA}

\newcommand{\acknowAC}{This research is part of the University of Toronto’s Acceleration Consortium, which receives funding from the CFREF-2022-00042 Canada First Research Excellence Fund. }

\usepackage{soul, color}
\newcommand{\Note}[1]{}
\renewcommand{\Note}[1]{#1}  

\title{Dynamic Execution Horizon Prediction for Chunk-based Robot Policies}

\author[1,2,\dagger]{Yuchi Zhao}
\author[3,\dagger]{Miroslav Bogdanovic}
\author[1]{Arjun Sohal}
\author[1]{Liyu Tao}
\author[3]{Kourosh Darvish}
\author[1,2,3,4,5,6,7,8,10]{\newline Al\'an Aspuru-Guzik}
\author[1,2]{ Florian Shkurti}
\author[9]{Animesh Garg}

\affiliation[1]{\addressCS}
\affiliation[2]{\addressVECTOR}
\affiliation[3]{\addressAC}
\affiliation[4]{\addressCHEM}
\affiliation[5]{\addressMSE}
\affiliation[6]{\addressCHEMENG}
\affiliation[7]{\addressMEDICALSCI}
\affiliation[8]{\addressCIFAR}
\affiliation[9]{\addressGT}
\affiliation[10]{\addressNVIDIA}

\contribution[\dagger]{Equal contribution}

\abstract{

Action chunking has become a standard design in modern robot policies, from diffusion/flow policies to vision-language-action models, where the policy predicts a sequence of actions and executes a fixed number of them instead of acting one step at a time. However, this paradigm relies on a key assumption: a fixed execution horizon. During chunk execution, the policy operates open-loop, which is particularly problematic for fine-grained manipulation tasks that require frequent replanning. In practice, the execution horizon is typically chosen through empirical tuning and is highly task-dependent. To this end, we propose Dynamic Execution Horizon Prediction (DEHP), an effective method that trains a lightweight execution-horizon prediction branch using online reinforcement learning while keeping the pretrained chunk policy completely frozen. This makes the method compatible with black-box chunk policies and isolates the effect of adapting the execution horizon from changes to the underlying action generator. Across our evaluations, DEHP improves the success rate of different high-precision and long-horizon manipulation tasks by a large margin. Our qualitative analysis further shows that DEHP predicts shorter execution horizons during fine-grained stages of the task and longer horizons during free-space motion. In this way, DEHP balances the efficiency of open-loop chunk execution with the reactivity of closed-loop single-step control. Project page: 
\href{https://dehp-chunking.github.io/}{dehp-chunking.github.io}

\vspace{1em}

}

\date{\today}
\correspondence{%
\raisebox{-0.01\baselineskip}{%
\begin{tabular}[t]{@{}l@{}}
\small\email{allan.zhao@utoronto.ca}\\[-0.8em]
\small\email{miroslav.bogdanovic@utoronto.ca}
\end{tabular}%
}%
}

\begin{document}

\maketitle



\clearpage
\begin{figure*}[h]
    \centering
    \includegraphics[width=\textwidth]{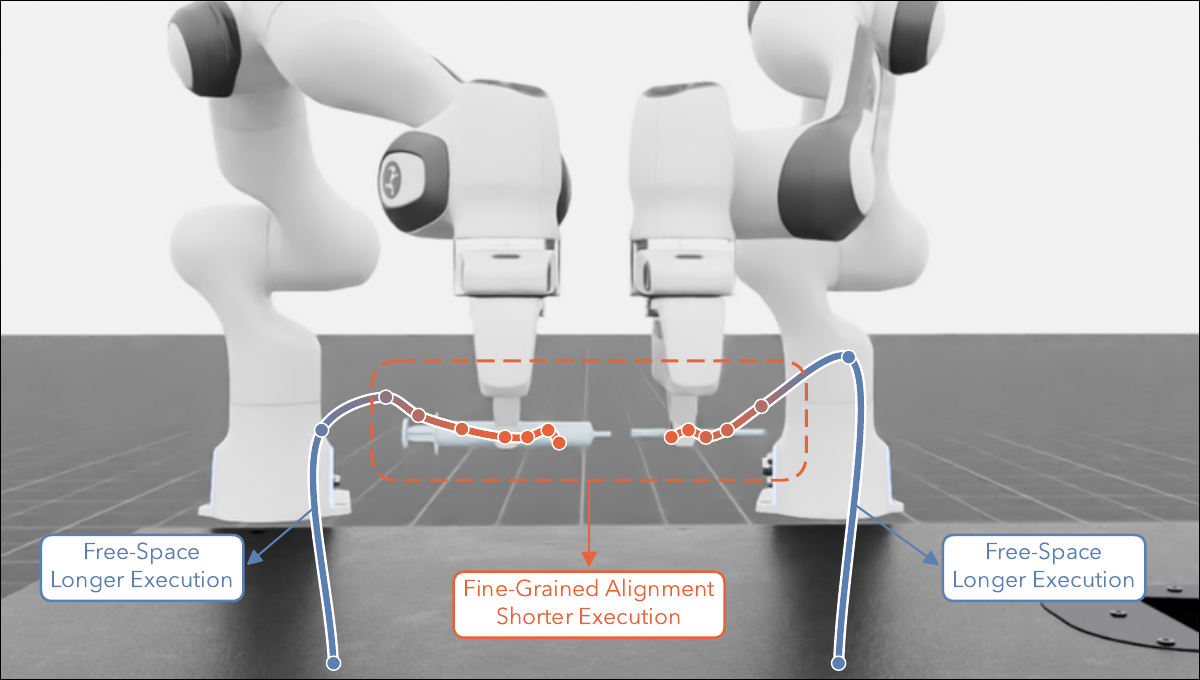}
    \caption{\textbf{Dynamic execution horizon in bimanual needle--syringe insertion.}
DEHP dynamically selects how many actions to execute from each predicted chunk based on the current observation and full action chunk. It uses longer horizons during free-space motion and shorter horizons during fine-grained alignment and insertion, enabling more frequent replanning when precision is required.}
    \label{fig:fig1}
\end{figure*}

\section{Introduction}

Robot policies have increasingly shifted from predicting single actions to generating full action chunks, both in behavior cloning (BC) and reinforcement learning (RL). Rather than predicting one action at a time, chunk-based policies predict a sequence of future actions and execute some part of that sequence before replanning. This improves temporal coherence and has led to strong performance in imitation learning, especially with diffusion- and flow-based policies~\cite{zhao2023learningfinegrainedbimanualmanipulation, chi2024diffusionpolicy}. However, existing chunk-based methods typically use a fixed prediction horizon and, more importantly, a fixed execution horizon throughout the task. This design creates a fundamental trade-off: longer execution horizons improve smoothness and efficiency but reduce reactivity, while shorter horizons increase responsiveness, but can hurt stability and temporal consistency.

We argue that execution horizons should be adaptive and change depending on the stage of the task. During free-space motion, a robot should be able to execute longer action sequences with little replanning, resulting in smooth and efficient behavior. On the other hand, in fine-grained manipulation, in particular contact-rich interactions, the robot should be able to replan more frequently in order to react to environmental feedback. A single fixed execution horizon cannot capture this variation across different task phases.

To this end, we propose Dynamic Execution Horizon Prediction (DEHP), a lightweight module that predicts the execution horizon conditioned on the current observation and the action chunk generated by a base policy.  We formulate dynamic execution horizon prediction as a semi-Markov decision process and optimize the horizon predictor using online RL with sparse binary rewards. Our results show that DEHP improves task success rates on both long-horizon and fine-grained manipulation tasks, even with a frozen base policy, and learns to adapt the execution horizon based on task progress. Our main contributions are: 

\begin{enumerate}[leftmargin=2.2em, itemsep=0.2em, topsep=0.1em, parsep=0pt]
    \item We propose Dynamic Execution Horizon Prediction (DEHP), which learns a categorical policy over the execution horizon $h \in \{1, \dots, H\}$ via online RL to maximize task return. DEHP is a small head on top of any pre-trained chunking policy, requires no architectural modification, and adds only a single forward pass at inference.     

    \item We provide a chunk-level PPO formulation for variable execution horizons, including a proof that the chunk-level discounted objective coincides with the base-MDP discounted return, along with a state-only critic and a step-discounted GAE estimator at chunk boundaries.

    \item We demonstrate that DEHP consistently improves success over tuned fixed-horizon baselines across assembly and fine-grained insertion tasks, and show that the learned horizons align with task phases that require different levels of reactivity.

\end{enumerate}

\section{Related Work}
\textbf{Action Chunking Policies}
Action chunking has become a common design choice in behavior cloning policies for robotic manipulation. Instead of predicting a single action at each step, chunk policies predict a sequence of H future actions and execute a subset of them before replanning. ACT\cite{zhao2023learningfinegrainedbimanualmanipulation}, the first work on action chunking, trains such policies with supervised regression and uses a Transformer to predict action chunks. Building on this idea, Diffusion Policy \cite{chi2024diffusionpolicy} models action chunks using a diffusion process, where the observation conditions an iterative denoising procedure from noisy actions to a clean action sequence, leading to strong improvements in success rate and multimodal action modeling. Most vision-language-action models \cite{octo_2023, black2026pi0visionlanguageactionflowmodel, intelligence2025pi05visionlanguageactionmodelopenworld, liu2025rdtb, lbmtri2025, gr00tn1_2025} also predict action chunks, using diffusion or flow matching for generating them. Prior work has also studied why action chunking is effective: Zhang et al.~\cite{zhang2025actionchunkingexploratorydata} provide both empirical and theoretical evidence that chunking improves control stability and mitigates compounding errors relative to single-step policies. However, there has been much less study of a key assumption underlying chunk-based policies which is the use of a fixed execution horizon. Our work directly addresses this gap.

\textbf{RL with Chunk-based Policies.}
Several recent works have incorporated RL to improve chunk-based diffusion and flow policies. DPPO~\cite{ren2025diffusion} formulates the diffusion denoising process as a Markov decision process (MDP) and directly fine-tunes the diffusion policy weights using Proximal Policy Optimization (PPO)~\cite{schulman2017proximalpolicyoptimizationalgorithms}. Other works explore value-based alternatives: Q-chunking~\cite{li2025reinforcement} and Decoupled Q-chunking~\cite{li2026decoupled} apply temporal-difference learning in a chunked action space to improve temporally coherent exploration and multi-step value estimation. Rather than updating the policy weights, DSRL~\cite{wagenmaker2025steering} steers a pretrained diffusion policy by learning in the latent noise space. On the other hand, residual RL methods keep the base policy frozen and train an MLP policy to predict corrective actions within each chunk. Methods such as ResiP~\cite{ankile2024imitationrefinementresidual} and PLD~\cite{xiao2025selfimprovingvisionlanguageactionmodelsdata} follow this paradigm to improve robustness without modifying the base policy. In contrast to these approaches, which refine or steer the generated actions under a fixed execution scheme, our method learns when to replan by dynamically adjusting the execution horizon based on task progress.

\textbf{Variable Execution Horizon.}
Concurrent work has also begun to study variable execution horizon selection. BID~\cite{liu2024bid} selects among sampled action chunks using temporal consistency and contrastive alignment. SGAC~\cite{so2025improving} determines the execution horizon from the cosine similarity between consecutive predicted chunks. TAS~\cite{weng2025temporalactionselectionaction} trains an action selector over temporally cached candidates, and MoH~\cite{Mixture_of_Horizons} fuses chunks sampled with different horizons through cross-horizon consensus. AAC~\cite{liang2024aac} and HiPolicy~\cite{zhang2026hipolicy} use inference-time entropy estimation as a proxy for uncertainty to adapt execution. These methods generally rely on sampling-based selection, auxiliary uncertainty proxies, or hand-crafted heuristics, which can increase inference cost or limit generalization. In contrast, DEHP learns an explicit execution horizon prediction branch from task return, enabling efficient adaptation without sampling or hand-crafted decision rules.

\section{Approach}
\begin{figure*}[t]
    \centering
    \includegraphics[width=\textwidth]{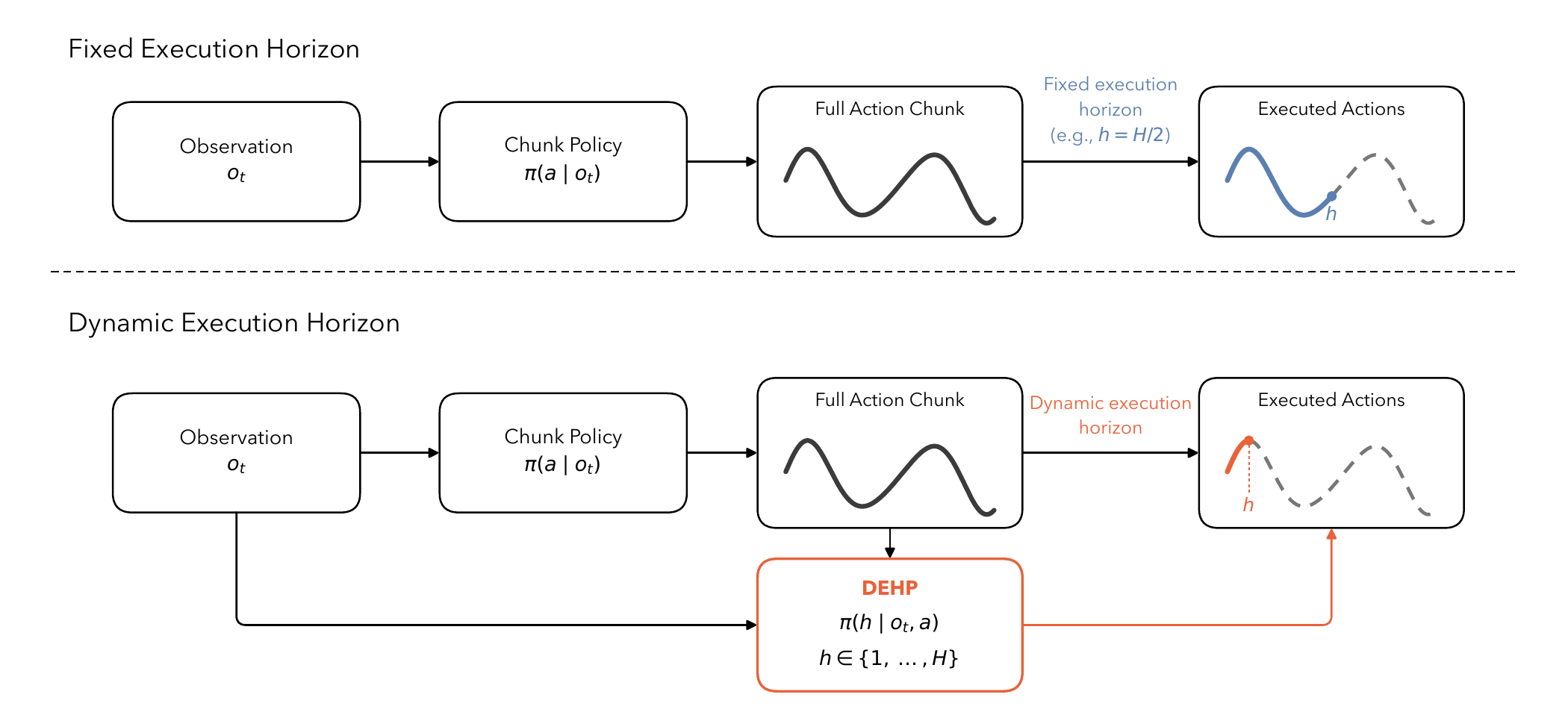}
    \caption{\textbf{Difference during inference between fixed execution horizon and our dynamic execution horizon.} \textbf{Top}: In the fixed execution horizon setting, the policy predicts a chunk of actions conditioned on the current observation and executes a predetermined, fixed number of actions before replanning. \textbf{Bottom}: In our method, the base policy similarly predicts a fixed-length action chunk; however, the DEHP module takes both the current observation and the predicted action chunk as input to dynamically determine the execution horizon. Only the selected subset of actions is executed before the next replanning step.}
    \label{fig:my_fig}
\end{figure*}

We aim to optimize a variable, state-conditioned execution horizon for a pretrained chunking policy. As shown in \autoref{fig:my_fig}, we add a lightweight horizon predictor on top of the base policy that, at each decision point, selects how many actions from the predicted chunk to execute before replanning. This allows the policy to commit to longer chunks during stable phases of a task and replan more frequently in phases that require high reactivity, without modifying the base policy. We train only the horizon predictor with online RL while keeping the base policy frozen.

\textbf{Problem Formulation.}
We consider a Markov decision process 
$\mathcal{M} = (\mathcal{S}, \mathcal{A}, P, R, \gamma)$, 
where $\mathcal{S}$ is the state space, $\mathcal{A}$ is the action space, 
$P$ denotes the transition dynamics, $R$ is the sparse binary reward function, and 
$\gamma \in (0,1)$ is the discount factor. We use the standard discounted-return objective 
$J(\pi) = \mathbb{E}_\pi[\sum_{t \geq 0} \gamma^t r_t]$. 
At each decision point, the base policy predicts a fixed-length action chunk 
$\mathbf{a}_{1:H} = (a_1, \dots, a_H)$. 
Our dynamic execution horizon prediction policy additionally selects an execution horizon 
$h \in \{1, \dots, H\}$, which determines how many actions from the predicted chunk are executed before replanning. 
We factor the joint distribution over the predicted chunk and execution horizon as
\begin{equation}
\pi(\mathbf{a}_{1:H},\, h \mid s) 
= 
\pi_{\mathrm{act}}(\mathbf{a}_{1:H} \mid s)\,
\pi_{\mathrm{len}}(h \mid s, \mathbf{a}_{1:H}),
\label{eq:factorization}
\end{equation}
where $\pi_{\mathrm{act}}$ is a pretrained base policy that predicts a length-$H$ action chunk at every decision point, and $\pi_{\mathrm{len}}$ is a categorical length head conditioned on the state and the full predicted chunk. This factorization has three benefits: (1) it makes DEHP a drop-in module for existing chunk policies; (2) it allows the horizon predictor to condition on the specific future behavior proposed by the base policy, and (3) it isolates the PPO update to a categorical length distribution when $\pi_{\mathrm{act}}$ is frozen. The first $h$ actions of $\mathbf{a}_{1:H}$ are executed open-loop, while the remaining $H-h$ actions are discarded before the next replanning step.

\textbf{Chunk-Level SMDP and Return Equivalence.}
Rolling out the chunking policy induces a sequence of chunk start times
$t_0 = 0$ and $t_{k+1} = t_k + h_k$, where $h_k \in \{1,\dots,H\}$ is the
execution horizon selected at chunk boundary $k$. The chunk-boundary process
$(s_{t_k})_{k\geq 0}$ evolves on the original state space $\mathcal{S}$, but
the time elapsed between consecutive decisions is itself variable and determined by the policy. We can therefore view the induced process as a
semi-Markov decision process (SMDP) over $\mathcal{S}$~\cite{sutton1999between}. For chunk $k$, we define the within-chunk discounted reward
\begin{equation}
\bar{R}_k := \sum_{j=0}^{h_k-1} \gamma^j r_{t_k+j},
\label{eq:chunk_reward}
\end{equation}
and the chunk-boundary value function
\begin{equation}
V^\pi(s)
:=
\mathbb{E}_\pi\!\left[
\sum_{k\geq 0} \gamma^{t_k} \bar{R}_k
\;\middle|\;
s_{t_0}=s
\right].
\label{eq:chunk_value}
\end{equation}
The discount between successive chunk-level decisions is therefore
$\gamma^{h_k}$, reflecting the number of environment steps executed before the next replanning event. A key question is whether optimizing this chunk-level objective changes the
underlying control problem. The following proposition shows that it does not.

\begin{proposition}[Return invariance]
For any chunking policy $\pi$ and any initial state $s$,
\begin{equation}
\mathbb{E}_\pi\!\left[
\sum_{k\geq 0} \gamma^{t_k} \bar{R}_k
\;\middle|\;
s_{t_0}=s
\right]
=
\mathbb{E}_\pi\!\left[
\sum_{t\geq 0} \gamma^t r_t
\;\middle|\;
s_0=s
\right].
\label{eq:return-invariance}
\end{equation}
\end{proposition}

Full details are provided in the appendix. The proposition is important for two reasons. First, it shows that the chunk-level formulation is not a surrogate objective: optimizing execution horizons at chunk boundaries is exactly equivalent to optimizing the original discounted return of the base MDP. Second, it implies that $V^\pi(s)$ in \autoref{eq:chunk_value} is simply the ordinary value of the step-level policy induced by $\pi$, evaluated at chunk boundaries. In particular, the value depends on the current state and current policy, not on the realized chunk index or execution horizon. As a result, in our on-policy setting a state-value critic $V_\phi(s)$ is sufficient, even though the horizon policy itself is conditioned on both the current state and the predicted action chunk. The critic therefore does not need to evaluate each candidate horizon separately. Since training is on-policy, $V^\pi(s)$ is the value under the current combined policy $\pi=(\pi_{\mathrm{act}},\pi_{\mathrm{len}})$, integrating over both predicted chunks and horizon samples. As $\pi_{\mathrm{len}}$ changes during training, the value function tracks the induced policy rather than maintaining horizon-specific values. This distinguishes our formulation from off-policy chunked-$Q$ methods~\cite{li2025reinforcement,li2026decoupled}, which must evaluate specific candidate chunks and therefore require action-conditioned critics. More discussion can be found in the appendix.

\textbf{Chunk-Level Advantage Estimation and PPO Objective.}
Given the SMDP formulation, the temporal-difference residual at chunk boundary \(k\) is
\begin{equation}
\delta_k
=
\bar{R}_k
+
\gamma^{h_k} V_\phi(s_{t_{k+1}})
-
V_\phi(s_{t_k}).
\label{eq:td_residual}
\end{equation}
We then estimate advantages using a step-discounted analogue of generalized advantage estimation (GAE):
\begin{equation}
\hat{A}_k
=
\sum_{\ell \geq 0}
(\gamma \lambda)^{t_{k+\ell}-t_k}
\delta_{k+\ell}.
\label{eq:gae}
\end{equation}
The exponent depends on elapsed environment steps rather than the number of chunk decisions, so future residuals decay according to how much environment time has passed. This estimator has the expected limiting behavior: when $\lambda=1$ and $V_\phi=V^\pi$, it is an unbiased estimate of the base-MDP advantage at chunk boundaries; when $h_k\equiv h$, it reduces to fixed-duration SMDP GAE with effective discount $\gamma^h$; and when $h_k\equiv 1$, it recovers ordinary step-level GAE.

Let $\theta$ denote the parameters of the horizon policy $\pi_{\mathrm{len}}$. Since the base chunk policy $\pi_{\mathrm{act}}$ is frozen, the per-chunk importance ratio reduces to
\begin{equation}
\rho_k(\theta)
=
\frac{
\pi_{\mathrm{len},\theta}(h_k \mid s_{t_k}, \mathbf{a}^{(k)}_{1:H})
}{
\pi_{\mathrm{len},\theta_{\mathrm{old}}}(h_k \mid s_{t_k}, \mathbf{a}^{(k)}_{1:H})
}.
\label{eq:ratio}
\end{equation}
We then optimize the standard clipped PPO surrogate
\begin{equation}
\mathcal{L}^{\mathrm{PPO}}(\theta)
=
-\mathbb{E}_k
\left[
\min\!\left(
\rho_k(\theta)\hat{A}_k,\;
\mathrm{clip}\!\left(\rho_k(\theta), 1-\epsilon, 1+\epsilon\right)\hat{A}_k
\right)
\right].
\label{eq:ppo_obj}
\end{equation}
Because $\pi_{\mathrm{len}}$ is categorical, $\log \rho_k$ is simply the difference between two log-probabilities from a softmax distribution, and can therefore be evaluated exactly and cheaply. As a result, the likelihood-estimation difficulties that arise when directly optimizing diffusion or flow-based action generators do not appear here: PPO is applied only to the lightweight horizon head, while the underlying chunk generator remains unchanged.

\textbf{Distributional Critic and Implementation Details.}
In our implementation, to improve value regression stability, the value critic is parameterized as a categorical distribution $p_\phi(\cdot \mid s) \in \Delta_B$ over a fixed support of $B$ atoms $v_1 < v_2 < \cdots < v_B$~\cite{bellemare2017distributional}. The scalar value used in the TD residual, GAE estimator, and return target is the mean of the predicted distribution, $V_\phi(s) = \sum_{b=1}^{B} v_b\, p_\phi(b \mid s)$.
Given the chunk-level return target
$V_{\phi_{\mathrm{old}}}(s_{t_k}) + \hat{A}_k$, we project this scalar target
onto the fixed atom support using a two-hot projection
scheme~\cite{bellemare2017distributional,imani2018improving,farebrother2024stop},
which assigns linearly interpolated mass to the two atoms that straddle the
target value. The critic is then trained by minimizing the cross-entropy between
the projected target distribution and the predicted categorical distribution.

For the horizon policy, we use a categorical predictor
$\pi_{\mathrm{len}}(h \mid s, \mathbf{a}_{1:H})$ over
$h \in \{1,\dots,H\}$. In the reported setup, the input to the horizon head is
the current observation together with the full predicted action chunk, implemented
as the concatenation of the current observation and the flattened action sequence.
This conditioning allows the model to select execution horizons based not only on
the current state, but also on the specific future behavior proposed by the base
chunk policy.

Finally, we initialize $\pi_{\mathrm{len}}$ with a uniform prior over
$\{1,\dots,H\}$ rather than warm-starting from a fixed horizon $h^\star$.
This avoids biasing the horizon policy toward a single execution length at the
start of training and allows DEHP to discover state-dependent horizon schedules
through online interaction.

\section{Experiments}
We conduct extensive experiments in challenging robotic environments, which demonstrate the effectiveness of dynamic execution horizon prediction for long-horizon and fine-grained manipulation tasks. Our evaluation is designed to answer the following questions:
\begin{enumerate}[leftmargin=2.2em, itemsep=0.2em, topsep=0.1em, parsep=0pt]
    \item How much can DEHP improve the success rate of chunk-based policies? 
    \item Can DEHP remain effective across different fine-grained and long-horizon manipulation tasks?
    \item What horizon schedules does DEHP learn, and why do they improve task performance?
\end{enumerate}

\textbf{Training Setup.} For each task, we first pretrain a chunk-based behavior cloning policy and then freeze its weights. DEHP is trained with online RL to predict the execution horizon, while the base policy remains fixed. We use Diffusion Policy as the base policy because of its strong performance and widespread use in robot imitation learning, and we use state observations in all experiments. For all environments, demonstrations are collected using a state machine with predefined waypoints tracked by an operational space controller (OSC). To increase data diversity, we randomize waypoint locations and object initial configurations, and inject action noise during data collection. We initialize the DEHP horizon head uniformly over the candidate execution horizons. Similar to DPPO, we warm-start the critic for several epochs before updating the execution horizon prediction head.

\textbf{FurnitureBench Assembly Environments.}
FurnitureBench~\cite{heo2023furniturebench} is a widely used benchmark for long-horizon manipulation and assembly. We evaluate DEHP on two FurnitureBench tasks, \texttt{one\_leg} and \texttt{round\_table}, which require skills such as picking, placing, insertion, and screwing. For each task, we design a task-specific state machine and collect 800 trajectories. The base policy observes the robot state and the states of all objects. Following the setup of ResiP~\cite{ankile2024imitationrefinementresidual}, the base policy predicts action chunks with prediction horizon 32.

\textbf{IsaacLab Insertion Environments.}
To evaluate DEHP on fine-grained manipulation, we construct two IsaacLab environments~\cite{mittal2025isaaclab}: multi-stage peg insertion and bimanual needle--syringe insertion. In multi-stage peg insertion, the robot sequentially picks up and inserts three pegs into holes with different clearances. In bimanual needle--syringe insertion, two robot arms separately grasp a needle and a syringe, then coordinate to assemble them in mid-air. For both tasks, we collect 1,000 successful trajectories and train the base diffusion policy with prediction horizon 16.

\subsection{How Much Do Dynamic Horizons Improve Success Rate?}

\begin{figure*}[h]
    \centering
    \includegraphics[width=\textwidth]{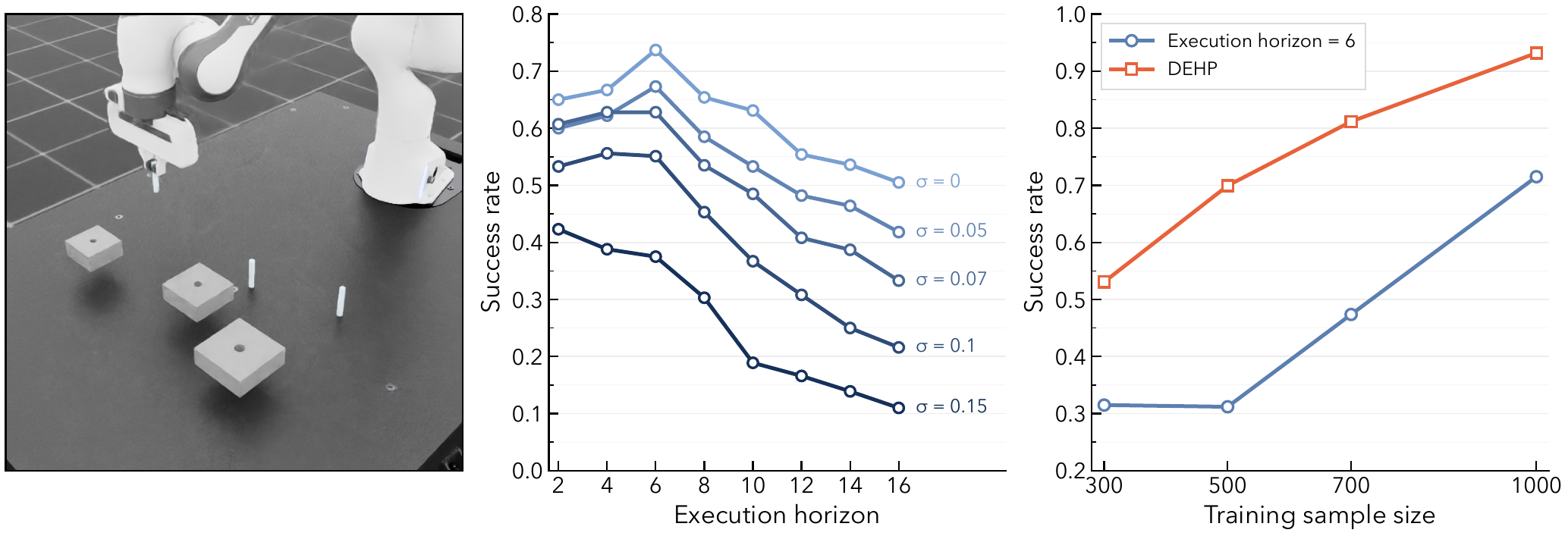}
    \caption{\textbf{Multi-stage insertion under different settings.}
\textbf{Left:} Three-stage insertion task, where the robot inserts the tightest, medium-fit, and loosest pegs in sequence.
\textbf{Middle:} Executing 6 actions performs best under low action noise, while shorter executions are better under higher noise.
\textbf{Right:} DEHP consistently improves performance across base policies trained with different numbers of demonstrations.}
    \label{fig:multi-insert-bc}
\end{figure*}

We first evaluate DEHP on the multi-stage insertion task in IsaacLab under different execution settings, as shown in \autoref{fig:multi-insert-bc}. In this task, the robot must sequentially pick up and insert three pegs into three holes. The insertion clearances are 2mm, 4mm, and 6mm for Stages 1, 2, and 3, respectively. To assess robustness to execution errors, we inject different levels of action noise during rollout and randomize object locations within a 7cm square region. For all evaluations, we report success rates over 1,000 environments, across 3 seeds. 

\textbf{Fixed-horizon BC.} 
As shown in \autoref{fig:multi-insert-bc}, the best fixed execution horizon depends on the execution-noise level. Under low action noise, executing 6 actions per chunk achieves the highest overall success rate. However, performance gradually degrades as the execution horizon increases, reflecting the reduced reactivity of longer open-loop execution. Under higher action noise, shorter horizons become more effective: at an action noise level of 0.15, executing 2 actions per chunk outperforms executing 6 actions, since more frequent replanning helps mitigate accumulated execution errors. Based on the low-noise setting, we use an execution horizon of 6 as the fixed-horizon BC baseline for the remaining evaluations in this section.

\textbf{DEHP.} 
We next evaluate DEHP using the same frozen base policy and report both per-stage and overall success rates. As shown in \autoref{tab:noise_stage_success}, DEHP achieves the highest overall success rate across all action-noise levels, improving overall success by 23.03\% on average over fixed-horizon BC. The largest gains occur in Stage 1, which has the tightest insertion clearance. This suggests that adaptive execution horizons are especially beneficial in the most precision-sensitive phase of the task, where frequent replanning allows the robot to correct small errors before they accumulate. As the action-noise level increases, DEHP continues to provide consistent improvements by dynamically adjusting the execution horizon during rollout.

\textbf{Effect of demonstration data scale.}
Finally, we study whether the gains from DEHP remain when the base BC policy is trained with more demonstrations. We train base policies with 300, 500, 700, and 1000 trajectories. As shown in \autoref{fig:multi-insert-bc}, increasing the amount of demonstration data generally improves fixed-horizon BC performance. However, DEHP consistently outperforms all fixed-horizon baselines across all dataset sizes. Even with the largest dataset, the best fixed-horizon BC policy remains below DEHP. This indicates that the performance gap is not only due to limited demonstration data, but also to the fixed execution-horizon assumption itself.

\begin{table}[!t]
    \centering
    \caption{\textbf{Success rate by insertion stage under different noise levels, averaged over 3 seeds.} DEHP achieves the highest success rate across all noise levels by adapting only the execution horizon, without fine-tuning the base policy. Overall success requires successful completion of all three insertion stages.}
    \label{tab:noise_stage_success}
    \small
    \setlength{\tabcolsep}{12pt} 
    \renewcommand{\arraystretch}{1.0}
    \begin{tabular}{c c c c c c}
    \toprule
    \multirow{2}{*}{\textbf{Action Noise Level}} & \multirow{2}{*}{\textbf{Method}} & \multicolumn{4}{c}{\textbf{Success Rate (\%)}} \\
    \cmidrule(lr){3-6}
     &  & \textbf{Stage 1} & \textbf{Stage 2} & \textbf{Stage 3} & \textbf{Overall} \\
    \midrule
    \multirow{2}{*}{0.00}
    & BC  & $77.90 \pm 1.92$ & $94.82 \pm 0.27$ & $96.80 \pm 0.53$ & $71.50 \pm 1.87$ \\
    & \cellcolor{gray!15}\textbf{DEHP} 
    & \cellcolor{gray!15}\textbf{\boldmath$96.57 \pm 0.81$} 
    & \cellcolor{gray!15}\textbf{\boldmath$98.31 \pm 0.71$} 
    & \cellcolor{gray!15}\textbf{\boldmath$98.14 \pm 0.71$} 
    & \cellcolor{gray!15}\textbf{\boldmath$93.17 \pm 1.27$} \\
    \midrule
    \multirow{2}{*}{0.05}
    & BC  & $74.53 \pm 1.95$ & $91.99 \pm 0.65$ & $95.67 \pm 0.27$ & $65.60 \pm 2.10$ \\
    & \cellcolor{gray!15}\textbf{DEHP} 
    & \cellcolor{gray!15}\textbf{\boldmath$93.90 \pm 0.50$} 
    & \cellcolor{gray!15}\textbf{\boldmath$96.91 \pm 0.39$} 
    & \cellcolor{gray!15}\textbf{\boldmath$97.22 \pm 1.04$} 
    & \cellcolor{gray!15}\textbf{\boldmath$88.47 \pm 1.16$} \\
    \midrule
    \multirow{2}{*}{0.07}
    & BC  & $72.20 \pm 0.35$ & $92.19 \pm 0.35$ & $93.14 \pm 0.71$ & $61.97 \pm 0.60$ \\
    & \cellcolor{gray!15}\textbf{DEHP} 
    & \cellcolor{gray!15}\textbf{\boldmath$91.93 \pm 1.08$} 
    & \cellcolor{gray!15}\textbf{\boldmath$95.97 \pm 0.62$} 
    & \cellcolor{gray!15}\textbf{\boldmath$96.26 \pm 0.26$} 
    & \cellcolor{gray!15}\textbf{\boldmath$84.93 \pm 1.17$} \\
    \midrule
    \multirow{2}{*}{0.10}
    & BC  & $66.87 \pm 0.85$ & $86.34 \pm 0.85$ & $92.02 \pm 1.37$ & $53.13 \pm 1.67$ \\
    & \cellcolor{gray!15}\textbf{DEHP} 
    & \cellcolor{gray!15}\textbf{\boldmath$86.27 \pm 0.26$} 
    & \cellcolor{gray!15}\textbf{\boldmath$94.43 \pm 0.71$} 
    & \cellcolor{gray!15}\textbf{\boldmath$94.61 \pm 1.12$} 
    & \cellcolor{gray!15}\textbf{\boldmath$77.07 \pm 0.84$} \\
    \midrule
    \multirow{2}{*}{0.15}
    & BC  & $58.70 \pm 0.26$ & $78.08 \pm 1.05$ & $82.32 \pm 1.85$ & $37.73 \pm 1.07$ \\
    & \cellcolor{gray!15}\textbf{DEHP} 
    & \cellcolor{gray!15}\textbf{\boldmath$76.90 \pm 1.57$} 
    & \cellcolor{gray!15}\textbf{\boldmath$90.59 \pm 0.79$} 
    & \cellcolor{gray!15}\textbf{\boldmath$88.19 \pm 1.32$} 
    & \cellcolor{gray!15}\textbf{\boldmath$61.43 \pm 1.55$} \\
    \bottomrule
    \end{tabular}
\end{table}

\subsection{Does DEHP Remain Effective Across Manipulation Tasks?}

We next evaluate whether DEHP generalizes beyond multi-stage insertion. We consider two FurnitureBench assembly tasks that require long-horizon execution and precise manipulation. In \textit{one-leg}, the robot picks up a table leg, inserts it into the tabletop, and screws it in place. The more challenging \textit{round-table} task additionally requires assembling the X-shaped table base. Following FurnitureBench, we report results under low and medium initial-object randomization.

\begin{figure*}[!t]
    \centering
    \includegraphics[width=\textwidth]{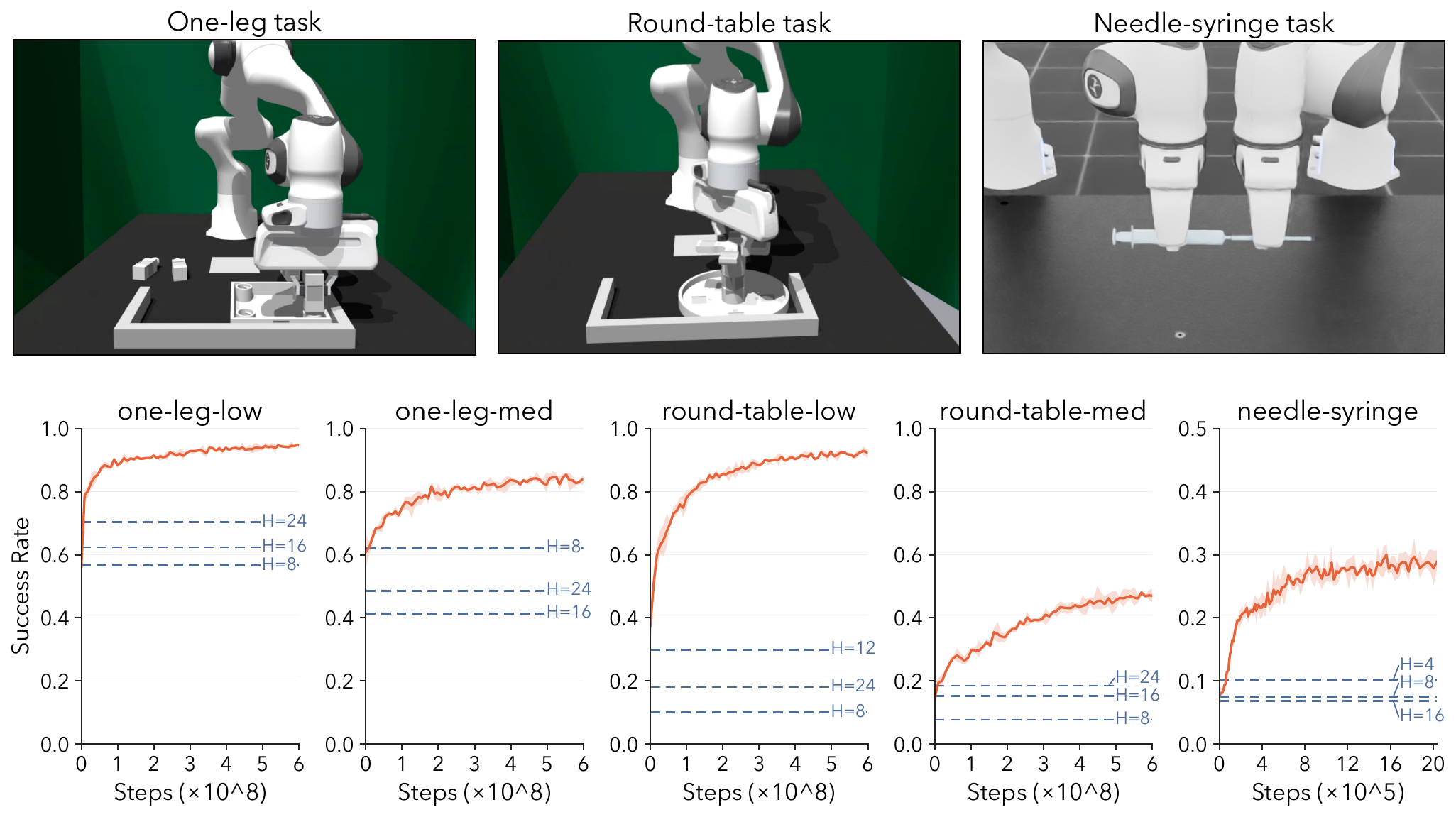}
    \caption{\textbf{Learning curves across manipulation tasks.}
Top: Example rollouts from the one-leg, round-table, and needle--syringe tasks.
Bottom: DEHP learning curves compared with fixed-horizon baselines across FurnitureBench assembly tasks and the IsaacLab bimanual insertion task. DEHP consistently improves over the best fixed execution horizon using the same frozen base policy.}

    \label{fig:success-learning-curve}
\end{figure*}

\autoref{fig:success-learning-curve} shows the learning curves of DEHP together with fixed-horizon baselines that execute a constant number of actions per chunk. On the \textit{one-leg} task, DEHP reaches a final success rate of 95.18\%, while the best fixed-horizon baseline achieves 70.30\% success with 24 executed actions per chunk. DEHP improves rapidly during the early phase of training and then increases more gradually. Because the horizon head is initialized uniformly, its initial performance can be lower than strong fixed-horizon baselines before it learns an effective adaptation strategy. The \textit{round-table} task shows the same trend but requires longer training due to greater complexity. DEHP improves success from 29.90\% to over 93.80\% under low randomization, and from 18.50\% to 53.03\% under medium randomization. These results show that dynamic execution horizons can provide large gains without changing the base policy, controller, or data-collection strategy.

An important finding is that \textit{one-leg} and \textit{round-table} favor different fixed execution horizons, despite using the same controller and similar state-machine demonstrations. This highlights a practical limitation of fixed-horizon execution: the best horizon must be tuned separately for each task. DEHP removes this manual tuning by adapting the execution horizon online.

Finally, we evaluate DEHP on a high-precision bimanual needle--syringe insertion task in IsaacLab. In this task, two robot arms separately grasp a needle and a syringe and then coordinate to assemble them in mid-air. The task requires precise alignment, with a 3mm needle opening radius and a 2mm syringe-tip radius. As shown in \autoref{fig:success-learning-curve}, the best fixed-horizon BC policy reaches only 10.20\% success. DEHP improves the same frozen base policy to 29.00\% success by dynamically adjusting the execution horizon. However, it also suggests that for precise tasks, performance may be further improved by adapting the execution horizon and fine-tuning the base policy.

\subsection{Why Do Dynamic Horizons Improve Performance?}
\begin{figure*}[t]
    \centering
    \includegraphics[width=1\textwidth]{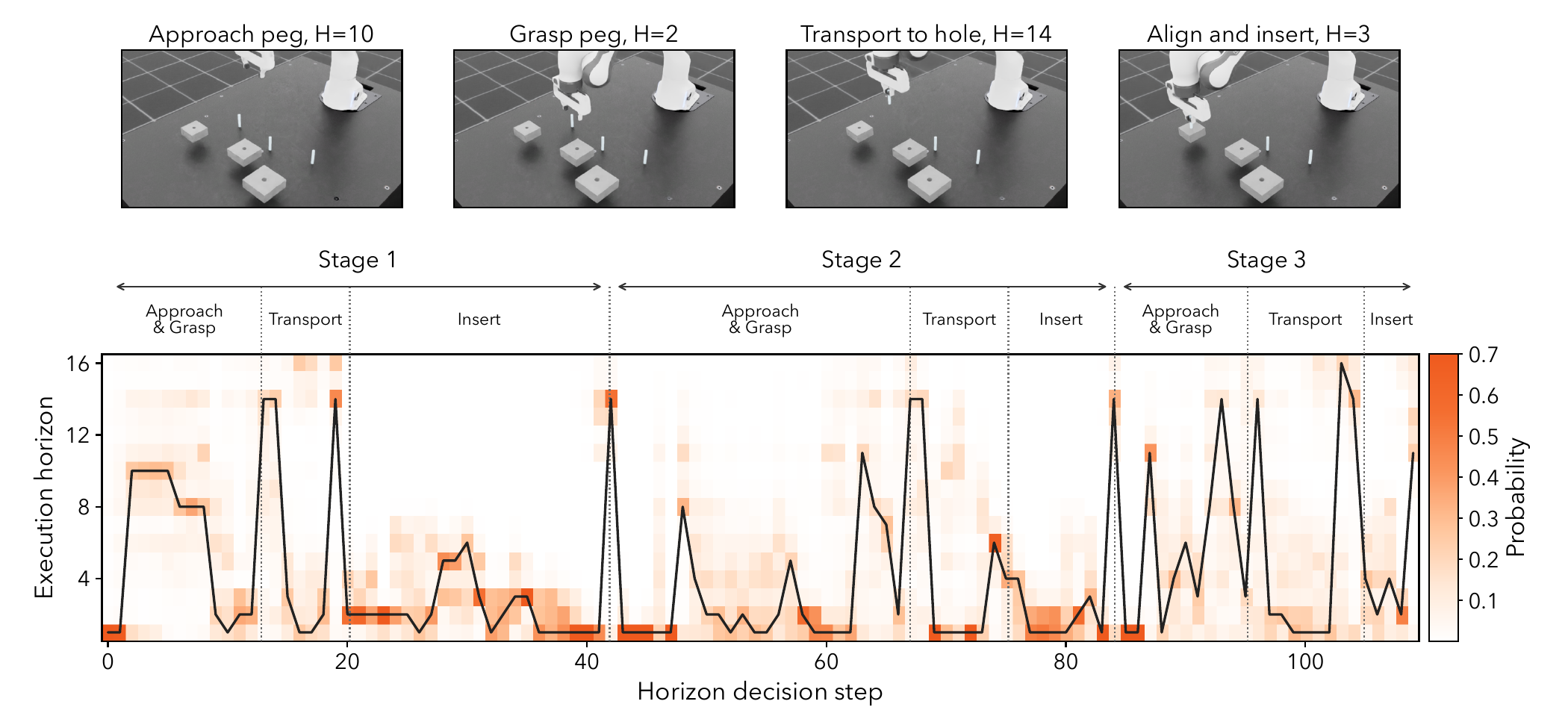}
    
    \caption{\textbf{Visualization of the DEHP output heatmap alongside keyframes from an episode of the multi-insertion task.} }
    \label{fig:hlm-heatmap}
\end{figure*}

To understand why dynamic horizons improve performance, we visualize DEHP's output distribution over an episode together with key task snapshots. As shown in \autoref{fig:hlm-heatmap}, DEHP learns a clear pattern in the three-stage insertion task: it predicts short horizons during feedback-sensitive phases, such as grasp alignment and insertion, and longer horizons during free-space motion, where more actions can be executed safely before replanning. Stage 1 is especially notable because it has the tightest insertion tolerance and requires the most action steps. During this stage, DEHP often predicts a horizon near 1, allowing the robot to replan almost every step and correct small errors during the most difficult insertion phase. Additional visualizations are provided in the Appendix.
\section{Conclusion}
In this work, we study execution horizons in action-chunking policies for robotic manipulation. We argue that the number of executed actions should adapt to task progress rather than remain fixed. We propose DEHP, a lightweight execution-horizon prediction module trained with PPO, which conditions on the current observation and predicted action chunk to decide when to replan while keeping the base behavior-cloning policy frozen. Across different manipulation tasks, DEHP consistently outperforms the best fixed-horizon baselines using the same base policy. These gains remain robust under action noise and persist as demonstration data increases, suggesting that fixed-horizon execution is a key limitation of chunk-based policies. Qualitative analysis shows that DEHP learns interpretable schedules, using longer horizons during free-space motion and shorter horizons during grasp alignment and insertion. Overall, adaptive execution horizons provide a simple and effective way to improve chunk-based robot policies without modifying the underlying policy weights.

\textbf{Limitations.}
In this work, we improve chunk-based robot policies by optimizing only the execution horizon, while keeping the pretrained base policy fixed. This design makes DEHP lightweight and preserves the behavior learned from demonstrations, but it also is limited by the capability of the base policy. A natural direction for future work is to combine dynamic execution horizon prediction with low-level policy fine-tuning, allowing DEHP to adapt when to replan while the low-level policy is updated to generate better action chunks. Such a combination may further improve task success. In addition, our experiments mainly focus on state-based policies for efficiency and controlled evaluation. However, DEHP treats the low-level chunk policy as a black box: it only uses the current observation and the predicted action chunk to decide the execution horizon. Although our experiments use state-based policies, DEHP is not inherently limited to them and could be extended to image-based policies.

\section{Acknowledgment}

A.A.-G. thanks Anders~G.~Fr{\o}seth for his generous support. A.A.-G. also acknowledges the generous support of Natural Resources Canada and the Canada 150 Research Chairs program. 
\acknowAC This research was enabled in part by compute resources provided by the Vector Institute and the Digital Research Alliance of Canada.

\clearpage


{
\small
\bibliography{references}
\bibliographystyle{assets/plainnat}
}


\clearpage

\appendix


\section{Appendix}
\subsection{Return invariance}
  \label{app:proofs}

  Let $\pi$ be a chunking policy with execution horizons $h_k\in\{1,\dots,H\}$, and
  let the chunk start times be $t_0=0$ and $t_{k+1}=t_k+h_k$. With the within-chunk
  discounted reward $\bar R_k:=\sum_{j=0}^{h_k-1}\gamma^{j}r_{t_k+j}$, the chunk-level
  discounted objective equals the base-MDP discounted return:
  \begin{align*}
  \mathbb{E}_\pi\!\left[\,\sum_{k\ge0}\gamma^{t_k}\bar R_k \;\middle|\; s_{t_0}=s\right]
  &= \mathbb{E}_\pi\!\left[\,\sum_{k\ge0}\gamma^{t_k}\sum_{j=0}^{h_k-1}\gamma^{j}r_{t_k+j} \;\middle|\; s_{t_0}=s\right]
  && \text{(definition of } \bar R_k\text{)}\\[4pt]
  &= \mathbb{E}_\pi\!\left[\,\sum_{k\ge0}\sum_{j=0}^{h_k-1}\gamma^{\,t_k+j}\,r_{t_k+j} \;\middle|\; s_{t_0}=s\right]
  && \\[4pt]
  &= \mathbb{E}_\pi\!\left[\,\sum_{t\ge0}\gamma^{t}r_t \;\middle|\; s_0=s\right].
  && \text{(}t_{k+1}=t_k+h_k\text{)}
  \end{align*}
  The last equality holds because chunk $k$ covers the consecutive timesteps
  $t_k,\dots,t_{k+1}-1$: since $t_{k+1}=t_k+h_k$, the chunks are contiguous and
  together account for every timestep exactly once. Consequently, the value function
  of the chunk-based policy, $V^\pi(s):=\mathbb{E}_\pi[\sum_{k\ge0}\gamma^{t_k}\bar R_k\mid s_{t_0}=s]$,
  is exactly the base-MDP value of the step-level behavior induced by $\pi$, measured
  at chunk boundaries. 

\subsection{Contrast with off-policy chunked-$Q$ approaches.}

Our use of a state-value critic contrasts with chunked $Q$-learning methods such as Q-chunking~\cite{li2025reinforcement} and Decoupled Q-Chunking (DQC)~\cite{li2026decoupled}, which operate in the offline or off-policy regime and therefore learn action-conditioned critics of the form $Q(s,\mathbf{a}_{1:h})$. In that setting, the critic must evaluate a specific candidate action sequence at state $s$, since the data-collection policy and the policy being evaluated may differ. When multiple execution horizons are allowed, this typically requires either a family of horizon-specific critics $\{Q^{(h)}\}_{h=1}^{H}$ or a horizon-augmented critic $Q(s,\mathbf{a}_{1:H},h)$ that is consistent across different horizons. DQC, for example, uses action-conditioned critics at different chunk lengths: a full-chunk critic for value backup and a distilled partial-chunk critic for policy extraction, together with an intermediate state-value function used for regression. In contrast, our on-policy formulation only requires estimating the value of states under the current policy. The critic therefore takes the simpler form $V_\phi(s)$ and does not need to evaluate arbitrary candidate chunks or maintain horizon-conditioned $Q$-functions. This simplicity comes at the cost of relying on on-policy data rather than directly leveraging offline datasets, but it removes the need for horizon-specific or horizon-augmented critic machinery.

\clearpage
\subsection{Additional Visualization of DEHP prediction on challenging bimanual task}
\begin{figure*}[h]
    \centering
    \includegraphics[width=0.9\textwidth]{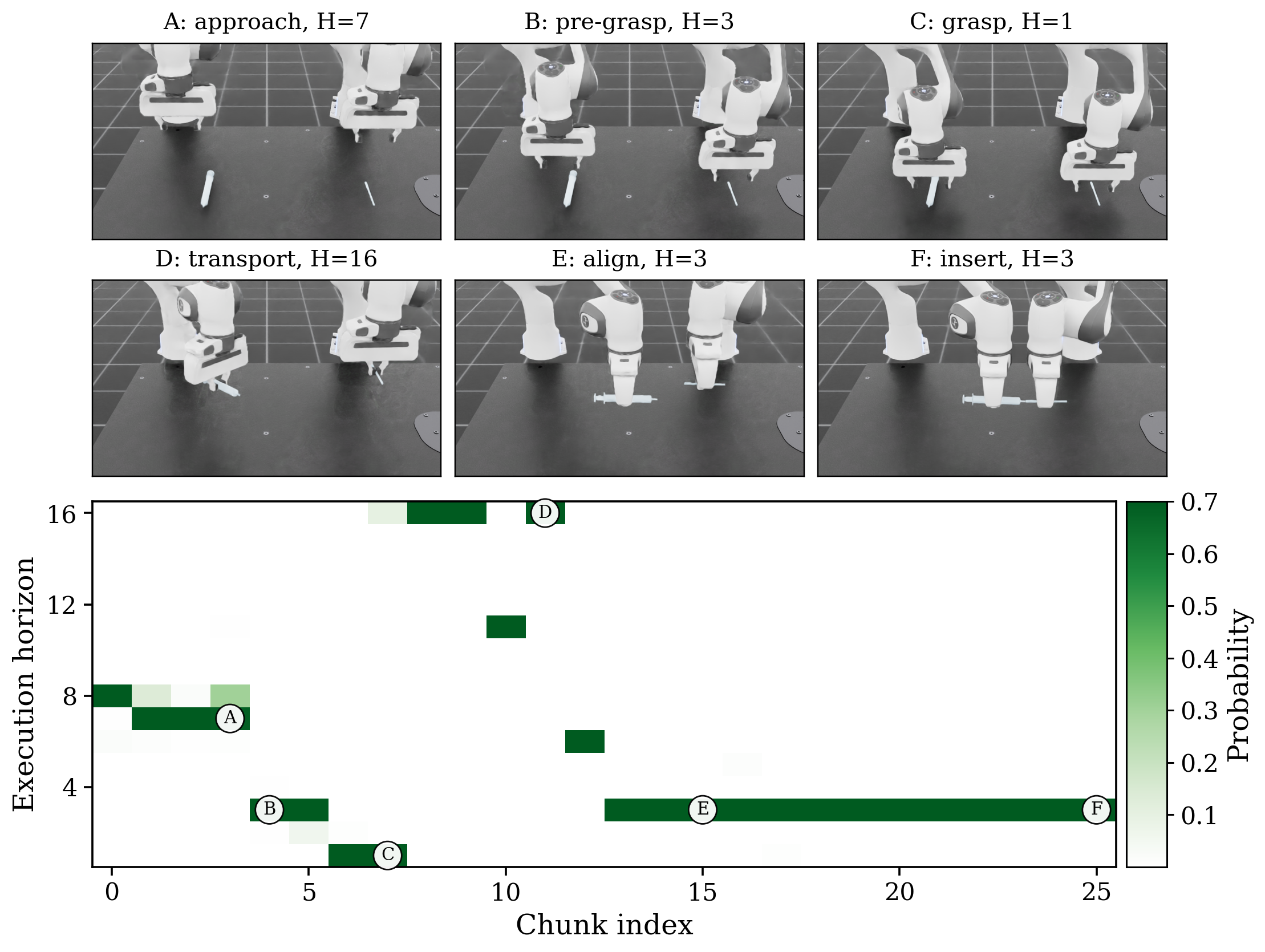}
    \caption{\textbf{Visualization of the DEHP output heatmap alongside keyframes from an episode of the bimanual needle–syringe insertion task.}}
    \label{fig:hlm-heatmap-needle}
\end{figure*}

Unlike the multi-insertion setting, this task requires tight coordination between two arms to assemble the needle and syringe in mid-air. As in \autoref{fig:hlm-heatmap-needle}, DEHP adapts its execution horizon accordingly, assigning longer horizons during coordinated approach motions and shortening them significantly during the final alignment and insertion, where small errors can easily lead to failure and rapid feedback is essential.

\clearpage
\subsection{Training Hyperparameters}

\begin{table*}[h]
\centering
\caption{Diffusion policy pretraining configuration. We report the task-specific hyperparameters used to train the frozen low-level chunk policy.}
\label{tab:diffusion_pretrain}
\small
\setlength{\tabcolsep}{8pt}
\renewcommand{\arraystretch}{1.50}
\begin{tabular}{lcccc}
\toprule
\textbf{Parameter} 
& \textbf{Multi-insertion} 
& \textbf{Needle insertion} 
& \textbf{One-leg} 
& \textbf{Round-table} \\
\midrule
\multicolumn{5}{l}{\textbf{Task and policy dimensions}} \\
Observation dimension & 44 & 28 & 58 & 44 \\
Action dimension & 7 & 14 & 10 & 10 \\
Prediction horizon $H$ & 16 & 16 & 32 & 32 \\
\midrule
\multicolumn{5}{l}{\textbf{Task-specific diffusion training}} \\
Batch size & 2048 & 2048 & 1024 & 2048 \\
Learning rate & $1{\times}10^{-4}$ & $1{\times}10^{-4}$ & $1{\times}10^{-4}$ & $1{\times}10^{-5}$ \\
\midrule
\multicolumn{5}{l}{\textbf{Task-specific network architecture}} \\
Policy hidden dimensions 
& $[1024,1024,1024]$ 
& $[1024,1024,1024]$ 
& $[1024]\times 7$ 
& $[1024]\times 7$ \\
Layer normalization & No & No & Yes & Yes \\
\midrule
\multicolumn{5}{l}{\textbf{Shared pretraining settings}} \\
\multicolumn{1}{l}{Denoising steps $T$} 
& \multicolumn{4}{c}{100} \\
\multicolumn{1}{l}{Weight decay} 
& \multicolumn{4}{c}{$1{\times}10^{-6}$} \\
\multicolumn{1}{l}{EMA decay} 
& \multicolumn{4}{c}{0.995} \\
\multicolumn{1}{l}{Policy backbone} 
& \multicolumn{4}{c}{MLP} \\
\multicolumn{1}{l}{Conditional MLP} 
& \multicolumn{4}{c}{$[512,64]$} \\
\multicolumn{1}{l}{Residual connections} 
& \multicolumn{4}{c}{Yes} \\
\bottomrule
\end{tabular}
\end{table*}

\begin{table*}[t]
\centering
\caption{DEHP fine-tuning configuration. The horizon-length model is trained on top of a frozen pretrained diffusion policy.}
\label{tab:dehp_finetune}
\small
\setlength{\tabcolsep}{8pt}
\renewcommand{\arraystretch}{1.50}
\begin{tabular}{lcccc}
\toprule
\textbf{Parameter} 
& \textbf{Multi-insertion} 
& \textbf{Needle insertion} 
& \textbf{One-leg} 
& \textbf{Round-table} \\
\midrule
\multicolumn{5}{l}{\textbf{Task-specific rollout setup}} \\
Parallel environments & 1000 & 1000 & 1000 & 1000 \\
Maximum episode steps & 640 & 200 & 1100 & 1100 \\
Diffusion horizon $H$ & 16 & 16 & 32 & 32 \\
DEHP horizon range & $1$--$16$ & $1$--$16$ & $1$--$32$ & $1$--$32$ \\
Rollout steps per iteration & 20 & 20 & 69 & 44 \\
Training iterations & 2000 & 2000 & 1000 & 1000 \\
\midrule
\multicolumn{5}{l}{\textbf{Task-specific optimization}} \\
Discount factor $\gamma$ & 0.9947 & 0.983 & 0.997 & 0.997 \\
GAE parameter $\lambda$ & 0.99 & 0.985 & 0.995 & 0.995 \\
PPO batch size & 500000 & 500000 & 17600 & 17600 \\
DEHP batch size & 2048 & 2048 & 4096 & 4096 \\
DEHP update epochs & 5 & 5 & 15 & 15 \\
\midrule
\multicolumn{5}{l}{\textbf{Shared fine-tuning settings}} \\
\multicolumn{1}{l}{Fine-tuning denoising steps $T_{\mathrm{ft}}$} 
& \multicolumn{4}{c}{5} \\
\multicolumn{1}{l}{Actor learning rate} 
& \multicolumn{4}{c}{$1{\times}10^{-5}$} \\
\multicolumn{1}{l}{Critic learning rate} 
& \multicolumn{4}{c}{$1{\times}10^{-3}$} \\
\multicolumn{1}{l}{DEHP learning rate} 
& \multicolumn{4}{c}{$1{\times}10^{-4}$} \\
\multicolumn{1}{l}{PPO update epochs} 
& \multicolumn{4}{c}{5} \\
\multicolumn{1}{l}{DEHP clip coefficient} 
& \multicolumn{4}{c}{0.05} \\
\multicolumn{1}{l}{Value critic} 
& \multicolumn{4}{c}{Distributional} \\
\multicolumn{1}{l}{Number of value atoms} 
& \multicolumn{4}{c}{201} \\
\multicolumn{1}{l}{Value support} 
& \multicolumn{4}{c}{$[0,1]$} \\
\bottomrule
\end{tabular}
\end{table*}


\end{document}